\documentclass[11pt,a4paper]{article}

\usepackage{times}
\usepackage{helvet}
\usepackage{courier}
\usepackage{algorithm}
\usepackage{tikz}
\usepackage{tikz-qtree}

\usepackage{multirow}
\usepackage{leftidx}
\usepackage{extarrows}
\usepackage{booktabs}
\usepackage{arydshln}
\usepackage{CJKutf8}
\usepackage{amsmath}
\usepackage{graphicx}
\usepackage{subfigure}
\usepackage[bottom]{footmisc}
\usepackage{amsfonts}
\usepackage{algorithmicx}
\usepackage{algpseudocode}
\usepackage{multirow, makecell,booktabs,multicol}
\usepackage{booktabs}

\usepackage[hyperref]{emnlp-ijcnlp-2019}
\usepackage{times}
\usepackage{latexsym}
\usepackage{url}
\usepackage{color}
\usepackage{makecell}

\aclfinalcopy 

\usepackage{cleveref}
\crefname{section}{§}{§§}
\Crefname{section}{§}{§§}
\newcommand{\MH}{\mathrm{MH}}
\newcommand{\RL}{\mathrm{RL}}
\newcommand{\F}{\mathrm{F}}

\algnewcommand\algorithmicinput{\textbf{Input:}}
\algnewcommand\INPUT{\item[\algorithmicinput]}
\algnewcommand\algorithmicoutput{\textbf{Output:}}
\algnewcommand\OUTPUT{\item[\algorithmicoutput]}

\newcommand{\ncite}[1]{\citeauthor{#1} (\citeyear{#1})}
\newcommand{\seq}[1]{\mathbf{#1}}

\newcommand{\argmax}{\operatornamewithlimits{\mathbf{argmax}}}

\DeclareMathAlphabet{\mathcal}{OMS}{cmsy}{m}{n}
\SetMathAlphabet{\mathcal}{bold}{OMS}{cmsy}{b}{n}

\title{Neural Machine Translation with Noisy Lexical Constraints}

\author{Huayang Li$^{1}${, }
  Guoping Huang$^{1}${, } 
  Deng Cai$^{2}${, }
  Lemao Liu$^{1}$ \\
  $^1$Tencent AI Lab{, } $^2$The Chinese University of Hong Kong \\
  {\tt \{alanili,donkeyhuang,redmondliu\}@tencent.com}\\
  {\tt thisisjcykcd@gmail.com}
  }

\date{}

\begin{document}
\maketitle
\begin{abstract}
 Lexically constrained decoding for machine translation has shown to be beneficial in previous studies. Unfortunately, constraints provided by users may contain mistakes in real-world situations. It is still an open question that how to manipulate these noisy constraints in such practical scenarios. We present a novel framework that treats constraints as external memories. In this soft manner, a mistaken constraint can be corrected. Experiments demonstrate that our approach can achieve substantial BLEU gains in handling noisy constraints. These results motivate us to apply the proposed approach on a new scenario where constraints are generated without the help of users. Experiments show that our approach can indeed improve the translation quality with the automatically generated constraints.

\end{abstract}

\section{Introduction}

Recently, a lot of efforts have been devoted to human-in-loop machine translation 
\cite{wuebker2016models,knowles2016neural,peris2017interactive,hokamp2017lexically,hasler2018neural,post2018fast}. In the simplest way, human can give hints on the words desired to show in the translation \cite{cheng2016primt}. To incorporate with this kind of lexical constrains, \ncite{hokamp2017lexically} and \ncite{post2018fast} proposed lexically constrained decoding for neural machine translation (NMT), which forces the inclusion of pre-specified words in the output by adding more complexity to standard beam search.

Lexically constrained decoding makes an implicit assumption that the given lexical constraints are perfect. However, in reality, people may make mistakes in their pre-specified constraints.  In our primary experiments, we found that there is a significant performance drop when moving such methods to noisy scenarios. Therefore, it is still an open question that how to improve the translation quality when constraints contain noises.

\begin{figure}
    \centering
    \includegraphics[width=6.5cm]{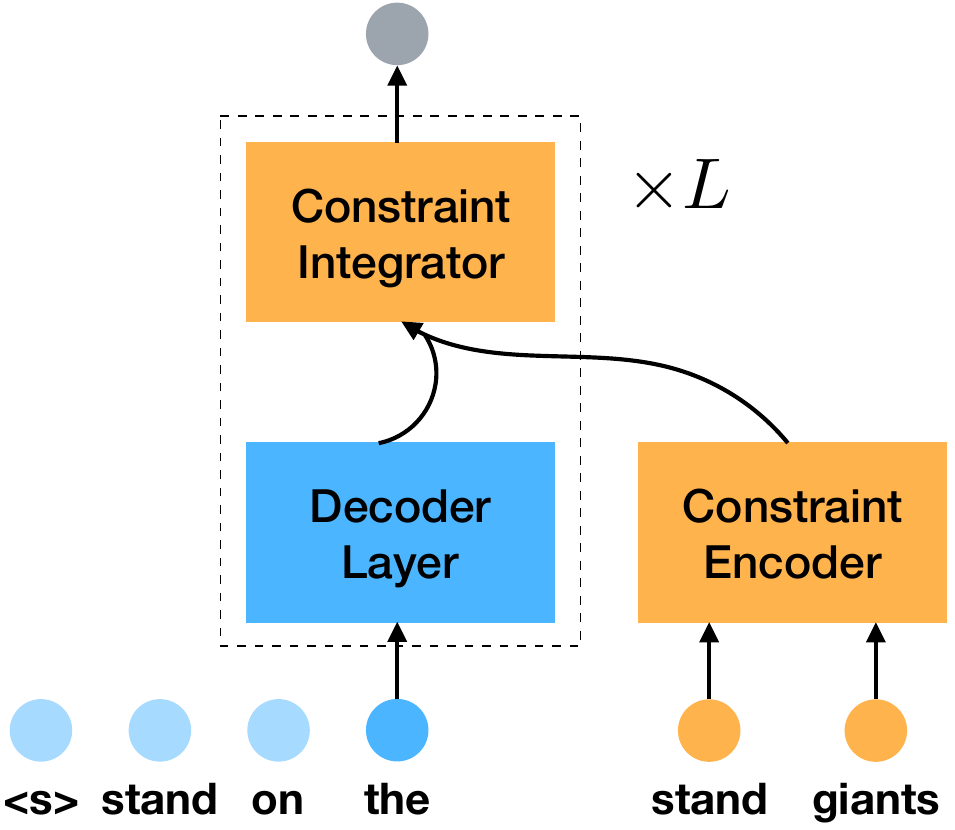}
    \caption{The decoder of the proposed framework.}
    \label{fig:struct}
\end{figure}
To our knowledge, this paper makes the first attempt to incorporate NMT with noisy constraints. In order to make better use of noisy constraints, we propose a novel framework to correct the noises as shown in Fig \ref{fig:struct}. Compared with standard Seq2Seq framework \cite{sutskever2014sequence,bahdanau+:2015}, our framework has two additional modules: the constraint encoder and the constraint integrator. The constraint encoder transforms lexical constraints into distributed representations. Specifically, two kinds of constraint encoder are proposed. The shallow one performs better when noisy rate is relatively low, while the deep one is more robust when the noises occur quite frequently. The integrator leverages constraints by treating them as external memories so that the mistaken constraints can be detected and fixed. Unlike \cite{hokamp2017lexically,post2018fast}, the additional overhead of our framework adds little time cost to the Seq2Seq framework. 


We test our methods with simulated correct constraints and randomly generated noisy constrains. Experiments show that our framework is able to improve the translation quality significantly even when the noisy rate of constraints is high. This motivates us to apply our framework to a new scenario where constraints are automatically generated. Experiments validate its usefulness.

We summarize our contributions as follows:
\begin{itemize}
\item
We make the first attempt to incorporate NMT with noisy constraints. A novel framework is proposed to handle the noisy constraints by leveraging them in a soft manner. Experiments show that this framework is \mbox{effective}.

\item Under the proposed framework, we systematically compare and analyse different models with the noisy constraints. The shallow constraint encoder is found to be more effective when noisy rate is relatively low, while the deep one is more robust with the high frequency of noises.

\item We propose to use automatically generated constraints for improving translation quality. Though the correctness of constraints is unguaranteed, our framework can still improve the translation quality.
\end{itemize}

\section{Background}
\subsection{Neural Machine Translation}
Suppose $\mathbf{x} = \left \langle x_{1}, \dots, x_{|\seq{x}|} \right \rangle$ is a source sentence with length $|\seq{x}|$ and $\mathbf{y} = \left \langle y_{1}, \dots, y_{|\seq{y}|} \right \rangle$ is the corresponding translation with length $|\seq{y}|$. 
Conventional NMT models compute the conditional probability $P\left( \mathbf{y} \mid \mathbf{x} \right)$ as follows: 
\begin{align}\label{eq:encdec}
P \left( \mathbf{y} \mid \mathbf{x};  \theta\right) & = \prod _{t} P \left( y_{t} \mid \mathbf{y}_{<t}, \mathbf{x};  \theta \right)  \notag\\ 
	& = \prod _{t}P(y_{t} \mid h_{t-1}^{d, L}),
\end{align}
where $\mathbf{y}_{<t}=\left \langle y_1, \dots, y_{t-1} \right \rangle$, $P(y_{t} \mid h_{t-1}^{d, L})$ is defined by a softmax function over the entire target vocabulary; $h_{i}^{d, L}$ is the decoding hidden vector at the last layer of NMT decoder ~\cite{bahdanau2014neural,vaswani+:2017}.

Most of the NMT models contain two standard modules: the encoder and the decoder. The encoder is used to represent the source sequence $\mathbf{x}$:
\begin{equation}
h_j^{e, l} = \varphi(\seq{h}^{e,l-1}, h_j^{e, l-1})
\label{eq:encoder}
\end{equation}
\noindent where $h^{e,l}_j$ refers to the hidden vector at the $l_{th}$ encoder layer, and particularly $h_j^{e, 1}$ is calculated from the word embedding and position encoding of $x_j$; $\seq{h}^{e,l}=\langle h_1^{e, l}, \cdots, h_{|\seq{x}|}^{e,l}\rangle$; $\varphi$ is a general 
encoder layer, for example, it is a bi-directional LSTM~\cite{bahdanau2014neural} or a self-attention based encoding network \cite{vaswani+:2017}. 

After encoding the source sequence, the decoder calculates the hidden vectors of target tokens as follows:
\begin{equation}
h^{d, l}_{i} = \phi(\seq{h}^{e,L}, \seq{h}^{d,l-1}_{<i+1}, h^{d, l-1}_i)
\label{eq:decoder}
\end{equation}
\noindent where $h_{i}^{d, l}$ indicates the decoding hidden vector at $l_{th}$ layer in the decoding phase, 
$\phi$ is a general decoder layer, which can be an attentional LSTM network~\cite{bahdanau2014neural} or a self-attention based decoding network~\cite{vaswani+:2017}.

In this paper, we employ the state-of-the-art Transformer as our baseline, which uses a self-attention based network for both encoder in Eq.~\eqref{eq:encoder} and decoder in Eq.~\eqref{eq:decoder}~\cite{vaswani+:2017}.



\subsection{Lexically Constrained Decoding}
\ncite{hokamp2017lexically} proposed an approach to improve the translation quality by using a sequence of lexical constraints $\seq{c}=\langle c_1, \cdots, c_{|\seq{c}|}\rangle$.~\footnote{Without loss of generality, in this paper, we assume that constraints consist of words rather than phrases as in \ncite{hokamp2017lexically}} The basic idea of their approach is to force the NMT model to output a translation which includes all elements in $\seq{c}$. Formally, it seeks to solve the following constrained optimization problem:
\begin{align}
 \label{eq:opt}
 & \argmax_{\seq{y}}  P(\seq{y} \mid \seq{x}) \notag\\
 \text{subject to} & \\
 & \sum_t \delta(c_i, y_t) > 0, \text{  } \forall i \notag
\end{align}
\noindent where $\delta(c_i, y_t)$ is 1 if $c_i = y_t$ and 0 otherwise. 

To solve this constrained optimization problem, \ncite{hokamp2017lexically} proposed a grid beam search algorithm. The grid beam search is applied to maintain the beam along two dimensions, where one is the length of the hypotheses and the other is the number of lexical constraints. Thus, its complexity is linear to the number of elements in $\seq{c}$. \ncite{post2018fast} improved the algorithm by dynamically allocating slots for different number of constraints in a beam. Though the complexity is independent on $|\seq{c}|$, it is still computationally expensive. In particular, both the mentioned methods assume that the constrained sequence requires critical criteria. Once the constraints contain some noises, the optimized translation is inclined to suffer from failures. 

\section{Methodology}

Since it’s difficult to guarantee that all the constraints are perfect in the real scenario, we assume that constraints $\mathbf{c}$ may contain some noises. In order to handle the noisy constraints, we propose a novel framework where a constraint encoder and a constraint integrator are added. In our framework, constraints are treated as external memories, which makes a great difference to existing approaches that use constraints in a hard manner \cite{wuebker2016models,knowles2016neural,hokamp2017lexically,post2018fast}.

\subsection{NMT with Constraint Memory}
 
Given $\seq{x}$, $\seq{c}$ and a prefix translation $\seq{y}_{<t}$, we generate a target word $\seq{y}_t$ according to
$P\left( {y}_t \mid \mathbf{x}, \seq{y}_{<t}, \seq{c}; \theta \right)$. 
Generally, our framework includes three components: 
it encodes $\seq{x}$ into a sequence of hidden vectors $\seq{h}^{e,L}$ in the same way as the NMT model
described in Section \S 2; it encodes the constraints $\seq{c}$ into another sequence of hidden vectors $E(\seq{c})$, 
i.e., the constraint memory; and it integrates the constraint memories into the decoder network to generate
the next token $y_t$. 

\subsubsection*{Constraint Memory Encoder}
We employ two different ways to encode the constraints $\seq{c}$ as continuous memories $E(\seq{c})$.



\textbf{Deep Encoder}:
The choice of deep encoder for constraints is identical to the encoder used for source sequence. As $\seq{c}$ includes tokens from the target vocabulary as $\seq{y}$, the
word embedding of this encoder is shared with the feeding word embedding of $\seq{y}$. This technique is found to be effective to avoid overfitting in our preliminary experiments. 

\textbf{Shallow Encoder}: 
Another choice is to employ the word embeddings to encode 
constraints $\seq{c}$. Despite its simplicity, experiments show that it works well especially when the noisy rate is relatively low. The shallow encoder has an additional advantage of removing redundant constraints. As the example illustrated in Fig \ref{fig:struct}, ``stand'' has been translated at timestep 2, 
then the ``stand'' in constraints will be redundant for timestep $t > 2$.
In this case, we can efficiently remove the redundant constraints in $\seq{c}$ by mask
technique. In contrast, it is inefficient for the deep encoder to 
dynamically mask out redundant constraints and recalculate the hidden units at each timestep.

\subsubsection*{Constraint Memory Integrator}


For the constraint memory integrator, we introduce two effective memory-augmented neural networks and propose a new one inspired by self-attention technique. Suppose $\hat{\seq{h}}^{d,l}_{<t}$ denotes decoding hidden vector sequence of $\seq{y}_{<t}$, we present three different methods to construct $\hat{h}^{d,l}_{t-1}$ and $P({y}_t\vert \seq{x}, \seq{y}_{<t}, \seq{c}; \theta)$ below.


\textbf{Gated Combination}: Following \citet{wang2017neural,tu2018learning,cao2018encoding,bapna2019non}, 
the decoding hidden vector $\hat{h}^{d, l}_{i}$ is derived with the constraint memories as follows:
\begin{equation}
\hat{h}^{d, l}_{i} = g_i * h + (1-g_i) * f_1(h, E(\seq{c}))
\label{eq:gatedhidden}
\end{equation}
\noindent where $h = \phi(\seq{h}^{e,L}, \hat{\seq{h}}^{d,l-1}_{<i+1}, \hat{h}^{d, l-1}_i)$, the gate $g_i = \textnormal{sigmoid}\circ f_2(h, E(\seq{c}))$, $f_1$ and $f_2$ are obtained by a general attention mechanism,~\footnote{As the length of $E({\seq{c}})$ is variant, we implement both $f_1$ and $f_2$ through an attention operator.} Then the integrator calculates the probability:
\begin{equation}
P\left( {y}_t \mid \mathbf{x}, \seq{y}_{<t}, \seq{c}; \theta \right)= P(y_{t}\mid \hat{h}^{d,L}_{t-1}).
\label{eq:genprob}
\end{equation}
\noindent where $\hat{h}_{t-1}^{d,L}$ denotes the decoding hidden vector at $L_{th}$ layer for the last timestep, which is the same as the NMT model in \S2. 

\textbf{CopyNet}: To generate a token from memories, \citet{gu2016incorporating} designed the CopyNet and \citet{feng-EtAl:2017:EMNLP2017} 
follow this idea to integrate the memory into NMT. 
It firstly introduces a constrained probability $P_c$, which is only defined within the tokens in memories, i.e. $\seq{c}$ in our scenario, as follows:
\begin{equation*}
 P_c(y_{t} \mid \hat{h}^{d,L}_{t-1}, E(\seq{c}))
\end{equation*}
\noindent where it is implemented by a 
constrained softmax over the tokens in $\seq{c}$. 
Unlike the gated combination method, we integrate the memory at the distribution level :
\begin{multline}
P\left( y_t \mid \mathbf{x}, \seq{y}_{<t}, \seq{c}; \theta \right)=  g_{t-1} * P (y_{t}\mid \hat{h}^{d,L}_{t-1}) +\\
					(1-g_{t-1} ) * P_c(y_{t} \mid \hat{h}^{d,L}_{t-1}, E(\seq{c}))
\end{multline}
\noindent where the gate $g_{t-1}$ is similar to that in Eq.~\eqref{eq:gatedhidden} but it is a scalar. To ensure the summation meaningful, $P_c(y_{t} \mid \hat{h}^{d,L}_{t-1}, E(\seq{c}))$ must be extended to the entire target vocabulary in advance, i.e. $P_c(y_{t} \vert \hat{h}^{d,L}_{t-1}, E(\seq{c}))=0$ if $y_t \neq c_i $ for all $i$.

\textbf{Self-Attention}: 
We propose a new memory integrator that is inspired by self-attention technique.
Formally, $\hat{h}_{i}^{d, l}$ is calculated by self-attention over the hidden representations of $\seq{y}_{<t}$ and memory $E(\seq{c})$ as follows:
\begin{multline}
h^{d,l}_{i} = \RL \circ \F \circ \RL \circ 
\MH  \Big( \\ \RL\circ \MH \big(\hat{h}_i^{d, l-1},  \seq{\hat{h}}_{< i+1}^{d,l-1}\Vert E(\seq{c})\big), \seq{h}^{e,L}\Big)
\label{eq:m1}
\end{multline}
\noindent where $\Vert$ denotes the concatenation of a pair of matrices, $\circ$ denotes the compositional function, 
$\RL$, $\F$ and $\MH$ are 
respectively residual, feed-forward, and multi-head attention sub-layers as described in \S2~\cite{vaswani+:2017}.
Then the probability $P({y}_t\vert \seq{x}, \seq{y}_{<t}, \seq{c}; \theta)$ is defined exactly the same as Eq.~\eqref{eq:genprob}.

\subsection{Training and Inference}

The above different encoders and integrators lead to six different models to compute $P(\seq{y} \vert \seq{x}, \seq{c}; \theta)$. Parameter $\theta$ is  optimized by the following objective:
\begin{equation*}
 - \sum_{\langle \seq{x}, \seq{r}, \seq{c}\rangle \in \mathcal{D}} \log P(\seq{r} \mid \seq{x}, \seq{c}; \theta)
\end{equation*} 
\noindent where $\mathcal{D} = \{\langle \seq{x}, \seq{r}, \seq{c}\rangle\}$ is a set of tuples of the source sentences, references and constraints. 
To train our models efficiently with stochastic gradient descent, we initialize it with the optimized baseline model and then fine-tune it on $\mathcal{D}$, which leads to the convergence within a few epochs in our experiments.

For inference, 
our decoding is an unconstrained optimization problem similar to 
the standard NMT:
\begin{equation}\label{eq:newopt}
\argmax_{\seq{y}} P(\seq{y}\mid \seq{x}, \seq{c}; \theta)
\end{equation} 
To approximately solve the above optimization problem, we employ the standard beam search algorithm which is similar to the decoding in our baseline model. Although we need additional overheads to handle the constraint memories, this is negligible to the time consuming of the decoding in NMT.
Thus, our search is as efficient as the search for conventional NMT.
In addition, unlike the lexically constrained decoding in Eq.\eqref{eq:opt}, our methods do not force a feasible $\seq{y}$ to include all words in $\seq{c}$. Hence, if the constraints in $\seq{c}$ include some noises, our methods have the potential to avoid copying these noises.


\section{Scenarios}
We consider three scenarios to apply our framework. The key to apply our framework in all scenarios is to generate constraints for each training pairs, which will be used to optimize the parameter $\theta$ in $P(\seq{y}\mid \seq{x}, \seq{c}; \theta)$.

\subsection{Scenario 1: Perfect Constraints}

In this scenario, we assume that constraints $\seq{c}$ for each test sentence $\seq{x}$ are perfect, where each $c_i$ is included in its ground-truth translation \cite{hokamp2017lexically}. 
This scenario is the same as that 
used by \citet{hokamp2017lexically} and \citet{post2018fast}. 

Different from the lexically constrained decoding methods in \citet{hokamp2017lexically} and \citet{post2018fast}, we require to extract constraints $\seq{c}$ for the training corpus. Following \ncite{post2018fast}, we simulate users' lexical constraints by taking words from the reference translation. In general, comparing to the high-frequency words, rare words are more difficult to appear in the translated sentences. Hence, we use the same method in \citet{grangier+auli:2018} by extracting the top-k rare words from the reference $\mathbf{r}$ as the perfect constraints.

\subsection{Scenario 2: Noisy Constraints}

This scenario has not been studied by previous research as described in \S1, in which we assume that each sentence has been provided a sequence of constraints with noises. 


We firstly generate a sequence of perfect constraints using the method in scenario 1 and then replace some of them with noises, which are assumed not included in the reference. The way to replacing perfect constraints is as follows: for each $c_i$, we replace it with a different yet similar word, named as \textit{replacing word}. For testing, we evaluate the performance with noisy constraints, using the replacing probability of 0.2, 0.4, 0.6, 0.8 and 1.0. While for training, it is set to 0.6.

To find the replacing word for each word,
a word-to-word mapping table is constructed by running GIZA++ \cite{och2003systematic} on the training corpus. Then we treat two target words $y_i$ and $y_j$ as the replacing words to each other if both key-value pairs $(x_k, y_i)$ and $(x_k, y_j)$ are in the translation table, where $x_k$ is a word in the source sentence \cite{ganitkevitch2013ppdb}.


\subsection{Scenario 3: Automatic Constraints}

In this scenario, we propose a new kind of constraints to improve the translation quality when there is no users' help. The constraints generation methods in the above scenarios are infeasible for this scenario because of the use of references.

As an alternative, we leverage a method to generate the constraints, where there is no guidance of references, as follows.
For each source sentence $\seq{x}$, we select the top-$k$ rare
source words from $\seq{x}$ for either the training set or the test set. Then we lookup the word-to-word translation table to obtain the best target word for each rare source word. The translation table is constructed as same as described in scenario 2.

\begin{algorithm}[t]
	\caption{Generic application of the proposed model}
	\begin{algorithmic}[1]
		   \INPUT
	       a bilingual training corpus $\mathcal{D}=\{\langle \seq{x}^i, \seq{r}^i\rangle\}$; a testing set $\mathcal{T}=\{ \seq{x}^j\}$
           \OUTPUT translation $\{\hat{\seq{y}}^j\}$
		\For{ $\langle\seq{x}^i, \seq{r}^i\rangle \in \mathcal{D}$}\Comment{get constraints for training}
		    \State generate $\seq{c}^i$ for $\langle\seq{x}^i, \seq{r}^i\rangle$
		\EndFor
		\State optimize $\theta$ according to Eq.\eqref{eq:newopt} over $\{\langle\seq{x}^i, \seq{r}^i, \seq{c}^i\rangle\}$\Comment{train model}
		\For{$\seq{x}^j$} \Comment{decoding process}
		\If {users provide $\seq{c}^j$ for $\seq{x}^j$}
		\State receive $\seq{c}^j$ from users
        \Else 
        \State generate $\seq{c}^j$ for $\seq{x}^j$
		\EndIf
		\State translate $\langle \seq{x}^j, \seq{c}^j \rangle$ to obtain $\hat{\seq{y}}^j$
		\EndFor
	\end{algorithmic}
\label{alg}
\end{algorithm}

The Algorithm \ref{alg} explains how to employ our approach for all the three different scenarios. 
In the lines 1 to 3, it generates the constraints for the training corpus, and then optimizes the model based on these constraints in line 4. In the lines between 5 and 12, it performs decoding for each test sentence before generating constraints if necessary. 
Specifically, for training, it employs the method designed for individual scenario to generate constraints in line 2. 
For testing (between line 5-12), it will either receive constraints from users for both scenarios 1 and 2 (line 7) or employ the method for scenario 3 which generates constraints (line 9) before the decoding in line 11.


\section{Experiments}
To validate our models, we conduct experiments on both the Chinese-to-English and French-to-English translation tasks. The translation quality is measured with case-sensitive BLEU-4 \cite{papineni2002bleu}.
 
\begin{table}[t]
\begin{center}
\begin{tabular}{l|c|c}
\toprule
Dataset &  Ch-En & Fr-En \\ \midrule
Train & 2004708 & 744228 \\
Dev & 2000 & 2655 \\
Test & 2000 & 2665 \\
\midrule
Averaged source length & 31.91 & 30.78\\
Averaged target length & 34.97 & 27.10 \\
\bottomrule
\end{tabular}
\end{center}
\caption{\label{table:dataset} Data statistics for both the Chinese-to-English (zh-en) and French-to-English (fr-en) translation tasks. }
\end{table}

\subsection{Datasets} 
The Chinese-to-English bilingual corpus includes news articles collected from several online news websites.~\footnote{These datasets will be publicly released soon.} After the standard preprocessing procedure as in \ncite{koehn2007moses},
we obtain about 2 million bilingual sentences in total. Then we randomly select 2000 sentences as the development and test datasets, respectively, and leave other sentences as the training dataset. 
The French-to-English bilingual corpus is from JRC-Acquis datasets \cite{steinberger2006jrc} and it is preprocessed
following \ncite{gu2017search}. This dataset is a collection of parallel legislative text of European Union 
Law applicable in the EU member states and thus it is a highly related corpus focusing on a specific domain.
The statistics information for both corpus is shown in Table \ref{table:dataset}.
In addition, we employ Byte Pair Encoding \cite{sennrich2015neural} on the previous datasets.
We maintain a source vocabulary of 33k tokens and a target vocabulary of 25k tokens on the Chinese-to-English task, and use 20k tokens for each of the vocabularies on the French-to-English task.

\subsection{Systems} 
We implement three baselines to compare with our models. 
The first one is the standard Transformer, which is an in-house implementation using Pytorch following
\ncite{klein2017opennmt}. The other two baselines are the lexically constrained decoding methods on top of the in-house Transformer: 
one is the grid beam search in \ncite{hokamp2017lexically} and the other is the lexically constrained decoding with dynamic beam allocation (an improved approach for the grid beam search) in \ncite{post2018fast} . These baselines are denoted by \textsc{Transformer}, \textsc{GBS} and \textsc{DBA}, respectively. To ensure the translation quality, we set the default beam size for GBS and \textsc{DBA} as suggested by \ncite{hokamp2017lexically} and \ncite{post2018fast}. The shallow and deep constraint encoder are referred as \textsc{SE} and \textsc{DE}. For the constraint integrator, we refer the gated combination as \textsc{Gate}, CopyNet as \textsc{Copy} and self-attention as \textsc{Attn}.

We employ the default hyper-parameters for all the systems. For the NMT models, We use 6 encoder layers and 6 decoder layers and set the size of word embedding to 512. To mitigate overfitting, the dropout technique is applied with dropout rate 0.1. Experiments show that manipulating self-attention on $L$ decoder layers is better, however using gated combination and CopyNet on the last decoder layer is more useful. We train all the models with Adam optimization algorithm and tune the number of iteration based on the performance of the development set. 

\subsection{Results} 
We consider three different translation scenarios (1-3) as presented in previous section, and for each scenario we test the introduced methods on Chinese-to-English and French-to-English task. 
Since it is costly to ask users to provide constraints for the test sentences on both Scenario 1 and Scenario 2, we simulate the user efforts by taking five rare words from the references following \ncite{hokamp2017lexically} and \ncite{post2018fast}. 
In particular, as presented in the previous section, we further add some noises to the constraints using the word-to-word table on Scenario 2. 

\subsubsection{Perfect Constraints} 

In this scenario, we compare the proposed model \textsc{SE-Attn}~\footnote{We implemented our methods with different settings as summarized in Table 2, and we found that shallow encoder works well when more constraints are perfect and thus we reported \textsc{SE-Attn} for comparison in this scenario.} with lexically constrained decoding (\textsc{GBS} and \textsc{DBA}) where all constraints are perfect.

\begin{table}[t]
\centering
\begin{tabular}{lccc} \toprule
{\quad\,\,\, Systems} & \multicolumn{1}{c}{Ch-En} & \multicolumn{1}{c}{Fr-En} & \multicolumn{1}{c}{Runtime} \\ \midrule
\textsc{Transformer} & 36.22 & 67.13 & 0.256\\
\quad + \textsc{GBS} & 42.17 & 70.99 & 5.692\\
\quad + \textsc{DBA} & 41.23 & 67.35 & 1.531\\
\quad + \textsc{SE-Attn} & 42.96 & 71.10 & 0.262\\ \bottomrule
\end{tabular}
\caption{\label{table:zh-sc2} BLEU and runtime comparison on perfect constraints for the Chinese-to-English and French-to-English tasks.
The runtime is measured by the time consumed by decoding one sentence on Chinese-to-English task.}
\end{table}

Table \ref {table:zh-sc2} depicts the comparison between \textsc{SE-Attn} and lexically constrained decoding in terms of both BLEU and decoding efficiency. Compared with \textsc{GBS} and \textsc{DBA}, \textsc{SE-Attn} gives a better performance in overall translation quality. The performance of \textsc{DBA} is slightly worse than \textsc{GBS} due to the aggressive pruning. In addition, it is shown that \textsc{SE-Attn} is comparable to \textsc{Transformer} in decoding efficiency, because it adds little time cost to the standard NMT framework.

\begin{table}[t]
\setlength{\tabcolsep}{1pt}
\centering
\scalebox{0.98}{
\begin{tabular}{lm{10mm}m{10mm}m{10mm}m{10mm}m{10mm}} \toprule
\multirow{2}{*}{\quad\,\,\, Systems} & \multicolumn{5}{c}{Number of the noises in constraints} \\ \cmidrule{2-6}
& \quad 1 &\quad 2 &\quad 3 &\quad 4 &\quad 5  \\  \midrule
\textsc{Transformer} &  36.22 & 36.22 & 36.22 & 36.22 & 36.22\\
\quad + \textsc{GBS} & 38.69 & 35.82 & 33.02 & 30.72 & 29.14 \\
\quad + \textsc{DE-Gate} & 38.23 & 38.28 & \textbf{38.26} & \textbf{38.05} & \textbf{37.90} \\
\quad + \textsc{SE-Gate} & 39.89 & 38.85 & 38.15 & 37.00 & 36.56 \\
\quad + \textsc{SE-Copy} & 39.23 & 38.25 & 37.74 & 36.74 & 35.89 \\
\quad + \textsc{SE-Attn} & \textbf{40.49} & \textbf{39.20} & 37.74 & 36.07 & 34.88 \\
\bottomrule
\end{tabular}}
\caption{BLEU comparison on noisy constraints
 for the Chinese-to-English task, where users totally provide 5 constraints with different number of noises.}
\label{table:zh-sc1}
\end{table}

\subsubsection{Noisy Constraints} \label{sec:noisy_results}

In the scenario of noisy constraints, we systematically compare several models and analyse the behaviors of them.
Firstly, we compare \textsc{GBS} with \textsc{Transformer}, which shows a performance drop of lexically constrained decoding. Then we go deep into the proposed framework and investigate the impact of different constraint encoders and integrators. 
Since there are six different settings for our models, it is time consuming to train all of them. Therefore, we first select from shallow and deep encoder methods using gated combination and then fix the encoder method to further compare different integrators. 

Table \ref{table:zh-sc1} shows the comparison of several systems on Chinese-to-English task, where constraints contain some noises. From this table, it can be seen that the performance of \textsc{GBS} degrades quickly as the number of the noises in constraints increases.  Compared with \textsc{Transformer}, \textsc{GBS} only achieves modest gains when 20\% of the constraints are noisy; but it completely fails if there are more than 2 noises in the constraints. On the contrary, all of our models under our framework are more robust than \textsc{GBS} for handling noisy constraints.

To analyse the behaviors of different constraint encoders and integrators, we compare several models under our framework. 
With the increasing number of noises, \textsc{SE-Gate} degrades much more slower than \text{DE-Gate}. This result shows that the deep constraint encoder is more robust when mistaken constraints occur frequently, while the shallow one is more superior when noisy rate of constraints is relatively low. Furthermore, based on the same constraint encoder, when more than 60\% constraints are perfect, \textsc{SE-Attn} performs the best compared with \textsc{SE-Gate} and \textsc{SE-Copy}. A possible reason is that, sharing the word embeddings with target words, SE could easily capture enough information of constraints. Notwithstanding DE has the potential to construct a better representation, the intricate architecture makes it hard to learn the parameters well.

\begin{table}[t]

\setlength{\tabcolsep}{1pt}
\centering
\scalebox{0.95}{
\begin{tabular}{lm{10mm}m{10mm}m{10mm}m{10mm}m{10mm}} \toprule
\multirow{2}{*}{\quad\,\,\, Systems} & \multicolumn{5}{c}{Number of the noises in constraints} \\ \cmidrule{2-6}
& \quad 1 &\quad 2 &\quad 3 &\quad 4 &\quad 5  \\  \midrule
\textsc{Transformer} & 67.13 & 67.13 & 67.13 & 67.13 & 67.13 \\
\quad + \textsc{GBS} & 67.01 & 64.05 & 61.57 & 59.65 & 57.82 \\
\quad + \textsc{DE-Gate} & 67.85 & 67.91 & 67.92 & 67.95 & \textbf{67.98} \\
\quad + \textsc{SE-Attn} & \textbf{69.38} & \textbf{68.94} & \textbf{68.53} & \textbf{68.10} & 67.59 \\ \bottomrule
\end{tabular}}
\caption{BLEU comparison on noisy constraints
 for the French-to-English task, where users totally provide 5 constraints including different number of noises.}
\label{table:fr-sc1}
\end{table}

\begin{table}[t]
\centering
\begin{tabular}{lcc} \toprule
\quad\,\,\, Systems & Ch-to-En & Fr-to-En \\ \midrule
\textsc{Transformer} & 36.22 & 67.13 \\
\quad + \textsc{GBS} & 35.14 & 67.43 \\
\quad + \textsc{DE-Gate} & 37.75 & 67.92 \\
\quad + \textsc{SE-Attn} & 37.66 & 68.46 \\
\bottomrule
\end{tabular}
\caption{\label{table:zh-sc3} BLEU comparison on automatic constraints for the Chinese-to-English and French-to-English task.}
\end{table}

Table \ref{table:fr-sc1} presents the comparison on French-to-English task.
From this table, we can see that the performance of \textsc{GBS} is very sensitive to the noisy rate of constraints.
In particular, \textsc{GBS} fails to improve the translation quality even if there is only one noise in the constraints. On the other hand, two models with the constraint memories are more stable as the number of noises increases. \textsc{DE-Gate} is unaffected by the noises number and \textsc{SE-Attn} achieves the gains up to 2.2 BLEU points with one noise, which coincides with the observations on the Chinese-to-English task.

\begin{CJK*}{UTF8}{gbsn}
\begin{table}[t]
    \centering
    \begin{tabular}{|l|m{5cm}|}\toprule
    Source & 他们 既 不 寻求 援助 , 也 不 祈望 重新 安置 。 \\
    Reference & They seek neither \textcolor{green}{assistance} nor resettlement . \\ 
    Constraint & \textcolor{red}{food} \\
    \textsc{GBS} & They did not seek \textcolor{red}{food} , nor did they wish to resettle .\\
    \textsc{SE-Attn} & They did not seek \textcolor{green}{assistance} , nor did they expect resettlement .\\ \midrule
    
    Source & 你 知道 的 我 在 日本 巡回演唱 \\
    Reference & I was on \textcolor{green}{tour} , in Japan , you know ? \\ 
    Constraint & \textcolor{red}{tourer} \\
    \textsc{GBS} &  You know , I 'm \textcolor{red}{tourer} in Japan\\
    \textsc{SE-Attn} & You know , I was on a \textcolor{green}{tour} in Japan . \\ \bottomrule
    \end{tabular} 
    \caption{Example translations with noisy constraints on Chinese-to-English task.}
    \label{tab:case}
\end{table}
\end{CJK*}

\subsubsection{Automatic Constraints}
Motivated by the above finding that our models are effective even if the noisy rate is very high, we propose to use automatically generated constraints for improving translation quality.
We conduct the experiments by leveraging automatic constraints and compare the performance of \textsc{DE-Gate} and \textsc{SE-Attn} with \textsc{Transformer}.

The results are shown in Table \ref{table:zh-sc3}. For Chinese-to-English task, it is observed that \textsc{DE-Gate} and \textsc{SE-Attn} outperform 
\textsc{Transformer} by a margin of 1.5 and 1.4 BLEU points. In a further comparison, we find the performance of \textsc{GBS} is even worse than the \textsc{Transformer} baseline. 
The main reason is that the noisy rate of the automatically generated constraints is up to $61.65\%$.
We also find a similar result on French-to-English task, which further demonstrates the usefulness of the proposed framework.


\subsection{Case Study}
Table \ref{tab:case} presents some examples of \textsc{GBS} and \textsc{SE-Attn}, which show that the proposed model could flexibly correct the noisy constraints. Given a random noisy constraint ``food'', \textsc{GBS} outputs a grammatically correct but semantic mismatched result. However, \textsc{SE-Attn} ignores the totally irrelevant constraint. As the second case shows, the concept of ``tour'' translated by \textsc{GBS} is unsatisfactory, while \textsc{SE-Attn} extracts
the semantic meaning of the noisy constraint `tourer` and generates a more appropriate substitute.

\section{Related Work}

In last few years, we have witnessed a notable revolution from statistical machine translation (SMT) \cite{koehn2003statistical,koehn2009statistical} to 
neural machine translation (NMT) \cite{sutskever2014sequence,bahdanau2014neural}. 
One of the vital appeals of NMT over SMT is that NMT is capable of modeling translation with sufficient global
information under an end-to-end framework \cite{bahdanau2014neural}. 
Though NMT generally exhibits decent translations, it still can not meet the strict requirements from users in real-world scenarios.

Constrained decoding provides an appealing solution to improve machine translation. It was pioneered by \ncite{foster2002text} where a SMT decoder sought to extend the manually prepared prefix \cite{barrachina2009statistical}. This idea was further explored by \ncite{Domingo2016Interactive}, which allowed users to provide
constraints at any position. 
\ncite{wuebker2016models} augmented the standard SMT model by introducing alignment information from a prefix and then trained the fine-grained SMT model for constrained decoding. 
Our approach is similar to \ncite{wuebker2016models} in the sense that we modify the model by integrating information from the constraints, but our model is on the top of NMT instead of the log-linear models in SMT. 

In the neural architecture, \ncite{wuebker2016models} and \ncite{knowles2016neural}
independently considered to extend prefixes with NMT models. 
\ncite{hokamp2017lexically} and \ncite{hasler2018neural} relaxed the constraints from the prefixes to general forms. \ncite{hokamp2017lexically} proposed a grid beam search algorithm organizing beams in terms of both the length of hypotheses and the number of constraints. \ncite{post2018fast} further improved it by using dynamical beams. 
Our approach is mostly close to both \ncite{hokamp2017lexically} and \ncite{post2018fast}, but we employ the constraints in a soft manner rather than a hard manner. In particular, different from all works mentioned above, our approach focuses on a general setting 
where constraints may include noises. 
Recently, it is interesting that \ncite{grangier+auli:2018} introduced a different kind of constraints (i.e. binary constraints) to improve NMT, but it employs these constraints in a soft manner as our approach. \ncite{dinu2019training} is contemporary to our work.

The proposed framework employs the constraints as external memories, and thus it is related to memory-based NMT which is widely studied by many researches such as~\citet{wang2017neural,feng-EtAl:2017:EMNLP2017,tu2018learning,zhang2016bridging,kaiser2017learning,cao2018encoding,bapna2019non}.
Under the proposed framework, we systematically compare several different memory integrators, which includes those from these researches and a new method inspired by self-attention. 
It is worth mentioning that the main focus of this paper is the framework which corrects noises in constraints in a soft manner rather than the memory integrators. In our scenario, we find that softly using constraints is superior to lexical constrained decoding, which is different from the findings in their scenarios. 

\section{Conclusion}
Lexically constrained decoding has shown to be helpful by previous works, but it depends on an ideal condition where lexical constraints provided by users are perfect. This paper makes the first
attempt to relax this condition to a more practical one and study NMT with noisy constraints. We propose a novel framework to augment the NMT model, which treats constraints as external memories. To demonstrate the effectiveness, it applies various models in different scenarios. Experiments show that the proposed framework can correct the noises in constraints. As a byproduct, we find that the shallow constraint encoder is more superior when noisy rate is relatively low, while the deep constraint encoder is more robust when the noises occur frequently. We propose a new scenario where constraints are generated without the help of users. Experiments show that our framework is capable of improving the translation quality with these automatically generated constraints.


\bibliography{emnlp-ijcnlp-2019}
\bibliographystyle{acl_natbib}

\end{document}